\documentclass[conference,letterpaper]{IEEEtran}
\IEEEoverridecommandlockouts

\usepackage{cite}
\usepackage{amsmath,amssymb,amsfonts}
\usepackage{algorithmic}
\usepackage{graphicx}
\usepackage{textcomp}
\usepackage{xcolor}
\usepackage{tikz}
\usepackage{pgfplots}
\usepackage{booktabs}
\usepackage{array}
\usepackage{listings}

\pgfplotsset{compat=1.18}
\usetikzlibrary{shapes,arrows,positioning,calc}

\usepackage[
  pdftex,
  hidelinks,
  bookmarks=false,
  pdfborder={0 0 0},
  colorlinks=false
]{hyperref}

\def\BibTeX{{\rm B\kern-.05em{\sc i\kern-.025em b}\kern-.08em
    T\kern-.1667em\lower.7ex\hbox{E}\kern-.125emX}}

\begin{document}
\sloppy

\title{\vspace{0.25in}Learning to Move Cities: Deep Meta-Models and Reinforcement Policies for Calibration and Control in Urban Networks}

\author{
\IEEEauthorblockN{Adewumi Augustine Adepitan}
\IEEEauthorblockA{\textit{Dept. of Civil, Environmental}\\
\textit{and Infrastructure Engineering}\\
\textit{George Mason University}\\
Fairfax, VA, USA\\
aadepita@gmu.edu}
\and
\IEEEauthorblockN{Christopher J. Haruna}
\IEEEauthorblockA{\textit{Dept. of Sustainability}\\
\textit{University of South Dakota}\\
Vermillion, SD, USA\\
christopher.haruna@coyotes.usd.edu}
\and
\IEEEauthorblockN{Oluwasegun Adegoke}
\IEEEauthorblockA{\textit{School of Information Studies}\\
\textit{Syracuse University}\\
Syracuse, NY, USA\\
oladegok@syr.edu}
\and
\IEEEauthorblockN{Ayooluwatomiwa Ajiboye}
\IEEEauthorblockA{\textit{Department of Computer Science}\\
\textit{George Mason University}\\
Fairfax, VA, USA\\
aajiboye@gmu.edu}
\and
\IEEEauthorblockN{Oluwatobi Oluwasakin}
\IEEEauthorblockA{\textit{Dept. of Transportation}\\
\textit{and Computing}\\
\textit{Federal University of Technology}\\
Akure, Nigeria\\
oluwasakintmt120653@futa.edu.ng}
}

\IEEEpubid{\makebox[\columnwidth]{979-8-3315-XXXX-X/26/\$31.00~\copyright2026 IEEE \hfill}
\hspace{\columnsep}\makebox[\columnwidth]{}}

\maketitle
\vspace{-0.15in}

\begin{abstract}
Urban transportation networks present complex optimization challenges spanning calibration of high-fidelity simulators and real-time operational control. This paper presents a shared latent-space framework that connects simulator calibration and reinforcement learning control through a common learned representation of urban traffic dynamics. First, we develop a combinatorial MLP-autoencoder architecture that learns low-dimensional manifolds linking simulator inputs (origin-destination demand, network parameters) to outputs (travel times, congestion patterns), enabling efficient Bayesian optimization for calibration. This approach demonstrates superior sample efficiency compared to traditional dimension reduction methods, achieving better fit to observational data within fixed computational budgets. Second, we implement a deep Q-learning agent with experience replay and target networks to optimize dynamic traffic assignment through scheduling and routing adjustments. In empirical evaluations on benchmark networks, our approach reduces system-wide travel times by up to 51\% compared to baseline operations. The learned latent representation is not only used to reduce the dimensionality of Bayesian calibration, but is also incorporated into the reinforcement learning state representation, allowing the control policy to operate on compressed and calibrated traffic dynamics. This shared latent-space formulation provides a unified pathway from simulator calibration to adaptive operational control within intelligent transportation systems. Our results highlight the transformative potential of deep learning methods in urban mobility planning and management, particularly for large-scale networks where traditional optimization approaches face computational bottlenecks.
\end{abstract}


\section{Introduction}

\IEEEPARstart{T}{he} growing complexity of urban transportation systems demands increasingly sophisticated computational approaches for both planning and operational tasks. Traditional methods face significant challenges in scaling to metropolitan-scale networks while maintaining accuracy and computational tractability \cite{nagel2012agent}. Two fundamental problems persist: the calibration of high-fidelity simulation models to match observed traffic patterns, and the optimization of dynamic control policies for improved network performance \cite{spiess1989optimal}.

Simulation-based transportation models have become essential tools for urban planning and intelligent transportation systems \cite{auld2016polaris}. These models, particularly agent-based approaches, capture complex interactions between travelers, infrastructure, and control systems \cite{sokolov2012flexible}. However, their utility depends critically on accurate calibration to real-world observations, a challenging inverse problem involving high-dimensional parameter spaces and computationally expensive model evaluations \cite{schultz2018bayesian}. Existing calibration methods, including Bayesian optimization and evolutionary algorithms, struggle with the curse of dimensionality when dealing with large urban networks \cite{hale2015optimization}.

Simultaneously, operational control of transportation networks requires adaptive policies that respond to dynamic conditions. Reinforcement learning offers a promising framework for such adaptive control \cite{sutton2018reinforcement}, but traditional approaches face limitations in handling the high-dimensional state and action spaces characteristic of urban networks \cite{abdullah2003reinforcement}. Recent advances in deep reinforcement learning, including double Q-learning, prioritized replay, and representation-aware policy learning, have demonstrated significant potential in complex control applications \cite{mnih2015human}, yet their application to transportation networks remains underexplored, particularly for integrated calibration and control.

This paper addresses both challenges through a shared latent-space framework that unifies simulator calibration and reinforcement learning control. Unlike existing approaches that treat calibration and operational control independently, the proposed framework learns a compressed transportation representation that supports both efficient Bayesian calibration and adaptive reinforcement learning-based network management. Our approach builds on recent work in simulation-based optimization \cite{chong2017simulation} and deep learning for transportation \cite{polson2017deep}, extending these methodologies to create a comprehensive solution for urban network management. The framework consists of two tightly connected components linked through a shared latent representation: a deep learning architecture for simulator calibration and a reinforcement learning system for dynamic network control. Unlike prior transportation learning frameworks that develop calibration and control pipelines independently, the proposed approach enables both modules to operate on a common compressed representation of transportation dynamics. The latent representation learned during calibration is reused as part of the reinforcement learning state space, enabling both modules to operate on a common compressed description of transportation network dynamics.

The main contribution of this paper is the development of a shared latent-space framework for urban transportation calibration and control. Unlike prior studies that treat simulator calibration and traffic control as separate tasks, the proposed framework learns a compact transportation representation from simulator input-output relationships and reuses this representation for both Bayesian calibration and reinforcement learning-based control.

Specifically, the framework employs a combinatorial MLP-autoencoder architecture to learn a low-dimensional latent representation of transportation simulator behavior. In the calibration stage, Bayesian optimization is performed within this latent space to improve sample efficiency and reduce computational cost. In the control stage, the same latent representation is incorporated into the reinforcement learning state description, enabling the DQN controller to operate on compressed and calibrated traffic dynamics rather than raw high-dimensional traffic states.

The remainder of this paper is organized as follows: Section II reviews relevant literature in transportation simulation, calibration methods, and reinforcement learning applications. Section III details our deep meta-model architecture for calibration. Section IV presents our reinforcement learning framework for network control. Section V describes experimental setup and results. Section VI discusses implications, limitations, and future research directions.

\section{Related Work}

\subsection{Transportation Simulation and Calibration}

Transportation simulation models have evolved from aggregate macroscopic approaches to detailed agent-based systems that capture individual traveler behavior \cite{nagel2012agent}. The POLARIS framework \cite{auld2016polaris} represents the state-of-the-art in large-scale agent-based transportation simulation, modeling individual trip chains and mode choices across metropolitan regions. However, the computational intensity of these models presents challenges for both calibration and operational use.

Calibration of transportation models typically involves adjusting input parameters to minimize discrepancy between simulated and observed outputs \cite{lu2015enhanced}. Traditional approaches include gradient-based methods \cite{cipriani2011gradient}, genetic algorithms \cite{ma2002genetic}, and simultaneous perturbation stochastic approximation \cite{lee2009new}. More recently, Bayesian optimization has emerged as a powerful framework for simulation calibration \cite{schultz2018bayesian}, leveraging Gaussian processes to model the objective function and guide parameter search efficiently.

The curse of dimensionality remains a fundamental challenge in calibration, as the parameter space grows exponentially with network size. Dimension reduction techniques, particularly active subspaces \cite{constantine2015active}, have been applied to identify important parameter directions. However, these linear methods may fail to capture complex nonlinear relationships in transportation systems. Deep learning approaches offer potential for more effective dimension reduction through their ability to learn nonlinear manifolds \cite{polson2017deep}.

\subsection{Deep Learning in Transportation}

Deep learning has demonstrated remarkable success in various transportation applications, including short-term traffic prediction \cite{polson2017deep}, spatio-temporal modeling, and network analysis applications \cite{wu2017framework}. The ability of deep neural networks to learn hierarchical representations from raw data makes them particularly suitable for complex transportation systems where traditional feature engineering is challenging.

Multi-layer perceptrons (MLPs) have been widely applied to transportation problems due to their universal approximation capabilities \cite{hornik1989multilayer}. Autoencoders, as unsupervised deep learning architectures, have shown promise in learning compressed representations of high-dimensional data \cite{hinton2006reducing}. Their application to transportation simulation calibration, however, remains largely unexplored, particularly in combination with MLPs for joint input-output modeling.

Recent work has begun to explore deep learning for simulation meta-modeling \cite{schultz2018bayesian}, but existing approaches typically focus on either dimension reduction or response prediction independently. Furthermore, these methods are generally limited to calibration tasks and do not consider how learned latent representations may support downstream operational control. In contrast, the proposed framework learns a shared latent transportation representation that is reused across both simulator calibration and reinforcement learning-based network control.

\subsection{Reinforcement Learning for Transportation Control}

Reinforcement learning provides a mathematical framework for sequential decision-making under uncertainty \cite{sutton2018reinforcement}. In transportation, RL has been applied to various control problems, including traffic signal timing \cite{abdullah2003reinforcement}, ramp metering \cite{belletti2017expert}, and vehicle routing \cite{larsson2016coordinated}.

Q-learning \cite{watkins1992q} represents a foundational RL algorithm that learns action-value functions through temporal difference learning. However, traditional Q-learning suffers from the curse of dimensionality when applied to large state-action spaces. Deep Q-networks (DQNs) \cite{mnih2015human} address this limitation by using deep neural networks to approximate Q-functions, enabling application to complex domains like video games and robotics.

In transportation, DQNs have shown promise for traffic signal control \cite{gershman2016using} and network routing \cite{wu2017framework}. However, most existing reinforcement learning approaches rely on raw traffic-state representations and are developed independently of simulator calibration processes. As a result, the learned policies often operate without leveraging structured representations of network dynamics learned during calibration. The proposed framework addresses this gap by integrating latent-space simulator representations directly into the reinforcement learning state formulation.

\begin{figure}[h]
    \centering
    \includegraphics[width=1.0\linewidth]{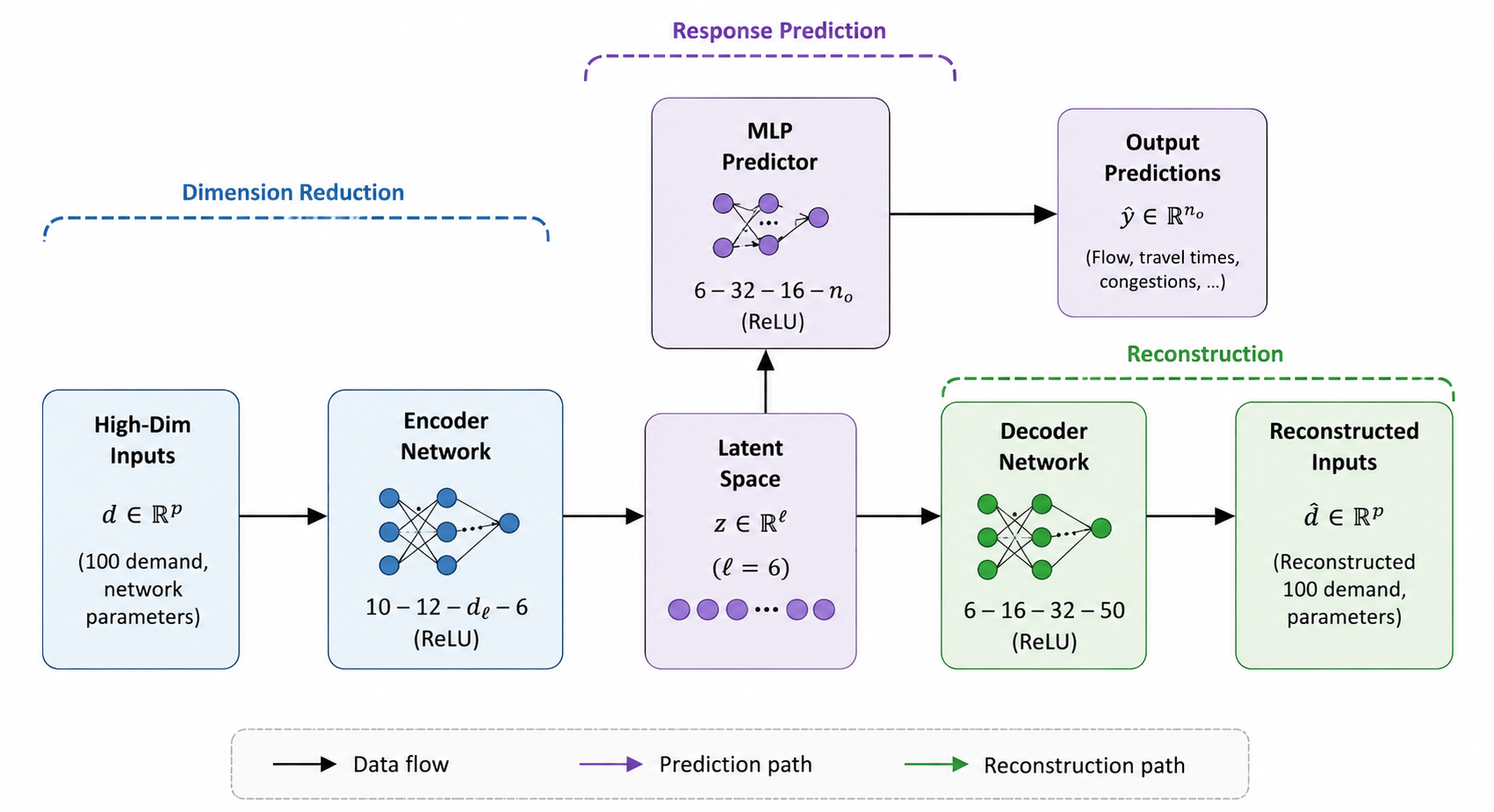}
    \caption{Combinatorial neural network architecture integrating autoencoder for dimension reduction with MLP output prediction}
    \label{fig:architecture}
\end{figure}
\section{Deep Meta-Models for Simulation Calibration}

The proposed framework integrates calibration and control through a shared latent-space representation of transportation network dynamics. Rather than treating calibration and reinforcement learning as independent modules, the framework first learns a compressed representation of simulator behavior using a combinatorial MLP-autoencoder architecture. This latent representation is subsequently reused by both the Bayesian calibration module and the reinforcement learning controller. As a result, the reinforcement learning agent operates on traffic-aware latent features that already encode important demand, congestion, and network-response patterns learned during calibration.
\subsection{Problem Formulation}

The calibration problem for transportation simulators can be formalized as an optimization task. Let $\theta \in \mathbb{R}^p$ represent the high-dimensional simulator input parameters, including origin-destination demand values, route choice parameters, link capacities, and traffic behavioral coefficients. Let $y \in \mathbb{R}^m$ denote simulator outputs, including link flows, average travel times, network delay, and congestion indicators collected across the network. The simulator implements a complex function $y = f(\theta)$ that is computationally expensive to evaluate.

Given observed data $y_{\text{obs}}$, the calibration objective is to find parameters $\theta^*$ that minimize the discrepancy between simulated and observed outputs:

\begin{equation}
\theta^* = \arg\min_{\theta \in \Theta} L(f(\theta), y_{\text{obs}})
\end{equation}

where $L$ is a loss function measuring simulation error, and $\Theta$ defines the feasible parameter space. The computational cost of evaluating $f$ makes direct optimization infeasible for large networks, necessitating efficient meta-model approaches.

\subsection{Combinatorial Neural Network Architecture}

Our approach employs a combinatorial architecture that integrates multi-layer perceptrons with autoencoders to address both dimension reduction and response prediction. As shown in Figure \ref{fig:architecture}, the network consists of three main components: an encoder network that maps high-dimensional inputs to a low-dimensional latent space, an MLP predictor that maps latent representations to simulation outputs, and a decoder network that reconstructs original inputs from latent representations.
The encoder, decoder, and MLP predictor were implemented as fully connected feedforward neural networks. The encoder architecture consisted of layers of size $50 \rightarrow 32 \rightarrow 16 \rightarrow 6$, where the six-dimensional latent vector represented the compressed transportation state. The decoder mirrored this structure using layers $6 \rightarrow 16 \rightarrow 32 \rightarrow 50$ to reconstruct simulator input parameters. The MLP predictor used a structure of $6 \rightarrow 32 \rightarrow 16 \rightarrow m$, where $m$ denotes the output dimension. Rectified Linear Unit (ReLU) activations were used in hidden layers, while tanh activation was applied to the latent layer to maintain bounded latent representations.

The encoder network implements a nonlinear dimension reduction:
\begin{equation}
z = g_{\text{enc}}(\theta; W_{\text{enc}}, b_{\text{enc}})
\end{equation}
where $z \in \mathbb{R}^d$ with $d \ll p$ represents the latent representation, and $g_{\text{enc}}$ is a deep neural network with parameters $W_{\text{enc}}, b_{\text{enc}}$.

The MLP predictor learns the input-output relationship in the latent space:
\begin{equation}
\hat{y} = g_{\text{mlp}}(z; W_{\text{mlp}}, b_{\text{mlp}})
\end{equation}

The decoder network ensures that the latent representation preserves essential information for input reconstruction:
\begin{equation}
\hat{\theta} = g_{\text{dec}}(z; W_{\text{dec}}, b_{\text{dec}})
\end{equation}

The complete architecture is trained end-to-end using a composite loss function:
\begin{equation}
\mathcal{L}_{\text{total}} = \lambda_1 \mathcal{L}_{\text{pred}}(y, \hat{y}) + \lambda_2 \mathcal{L}_{\text{recon}}(\theta, \hat{\theta}) + \lambda_3 \mathcal{L}_{\text{reg}}
\end{equation}
where $\mathcal{L}_{\text{pred}}$ measures prediction error, $\mathcal{L}_{\text{recon}}$ measures reconstruction error, $\mathcal{L}_{\text{reg}}$ provides regularization, and $\lambda_i$ are weighting coefficients.

\subsection{Bayesian Optimization in Latent Space}

The learned latent space enables efficient Bayesian optimization for calibration. Rather than searching in the original high-dimensional parameter space, optimization proceeds in the reduced latent space:

\begin{equation}
z^* = \arg\min_{z \in \mathcal{Z}} L(g_{\text{mlp}}(z), y_{\text{obs}})
\end{equation}

where $\mathcal{Z}$ is the latent space, typically defined as a hypercube based on the range of training data projections.

We employ Gaussian process (GP) regression to model the objective function in the latent space:
\begin{equation}
J(z) \sim \mathcal{GP}(\mu(z), k(z, z'))
\end{equation}
where $\mu(z)$ is the mean function and $k(z, z')$ is the covariance kernel. The GP posterior distribution guides the selection of evaluation points using acquisition functions such as expected improvement \cite{jones1998efficient}.

Once an optimal latent point $z^*$ is identified, the decoder network reconstructs the corresponding parameter values:
\begin{equation}
\theta^* = g_{\text{dec}}(z^*)
\end{equation}

This approach combines the sample efficiency of Bayesian optimization with the dimension reduction capabilities of deep learning, enabling effective calibration of high-dimensional simulators.

\subsection{Training Methodology}

The combinatorial network is trained using simulated data generated by running the transportation simulator with diverse parameter settings. We employ Latin hypercube sampling \cite{mckay2000comparison} to ensure good coverage of the parameter space.

Training proceeds in two phases. First, the autoencoder components (encoder and decoder) are pre-trained to learn effective latent representations using reconstruction loss:
\begin{equation}
\mathcal{L}_{\text{recon}} = \frac{1}{N} \sum_{i=1}^N \|\theta_i - \hat{\theta}_i\|^2_2
\end{equation}

Second, the complete network is fine-tuned using the composite loss function. Training was performed using the Adam optimizer with a learning rate of $0.001$, a batch size of 64, and a maximum of 200 epochs. Early stopping with patience of 15 epochs was applied based on validation loss to reduce overfitting. The bounded tanh activation function ensures that latent representations remain within a predictable range, facilitating subsequent optimization.

\section{Reinforcement Learning for Network Control}

\subsection{Problem Formulation}

The network control problem addresses dynamic decision-making in transportation systems. We formulate this as a Markov decision process (MDP) with state space $\mathcal{S}$, action space $\mathcal{A}$, transition dynamics $\mathcal{P}$, and reward function $\mathcal{R}$.

The state $s_t \in \mathcal{S}$ captures relevant network conditions at time $t$, including current demand, accumulated delays, and network occupancy. The action $a_t \in \mathcal{A}$ represents control decisions, such as routing recommendations or scheduling adjustments. The reward $r_t = \mathcal{R}(s_t, a_t)$ quantifies immediate network performance and is defined as a weighted combination of total system travel time, average network delay, queue overflow penalties, and infeasible routing penalties. This formulation encourages the controller to reduce congestion while discouraging actions that violate network feasibility or create excessive queue accumulation.

The objective is to learn a policy $\pi: \mathcal{S} \rightarrow \mathcal{A}$ that maximizes expected cumulative reward:
\begin{equation}
J(\pi) = \mathbb{E}\left[\sum_{t=0}^\infty \gamma^t r_t \mid \pi\right]
\end{equation}
where $\gamma \in [0,1]$ is a discount factor balancing immediate and future rewards.

\subsection{Deep Q-Learning Framework}

We employ deep Q-learning to learn optimal control policies. The Q-function $Q^\pi(s,a)$ represents the expected cumulative reward when taking action $a$ in state $s$ and following policy $\pi$ thereafter:
\begin{equation}
Q^\pi(s,a) = \mathbb{E}\left[\sum_{\tau=t}^\infty \gamma^{\tau-t} r_\tau \mid s_t=s, a_t=a, \pi\right]
\end{equation}

The optimal Q-function satisfies the Bellman equation:
\begin{equation}
Q^*(s,a) = \mathbb{E}\left[r + \gamma \max_{a'} Q^*(s',a') \mid s,a\right]
\end{equation}

Deep Q-networks (DQNs) approximate $Q^*(s,a)$ using a neural network $Q(s,a;\theta)$ parameterized by $\theta$. The network is trained by minimizing the temporal difference error:
\begin{equation}
\mathcal{L}(\theta) = \mathbb{E}_{(s,a,r,s')} \left[\left(r + \gamma \max_{a'} Q(s',a';\theta^-) - Q(s,a;\theta)\right)^2\right]
\end{equation}

where $\theta^-$ are parameters of a target network that is periodically updated to improve training stability.

\subsection{Network Architecture and Training}

Our DQN architecture processes state information using fully connected layers of size $64 \rightarrow 128 \rightarrow 64 \rightarrow |\mathcal{A}|$, where $|\mathcal{A}|$ denotes the number of control actions. ReLU activations were applied in hidden layers, and linear activation was used at the output layer to estimate action-value functions. The input state included both conventional traffic state variables and the learned latent transportation representation obtained from the calibration module. The state representation includes both current network conditions and historical patterns to capture temporal dependencies.

We implement several enhancements to improve training efficiency and stability:

\textbf{Experience Replay:} Transitions $(s_t, a_t, r_t, s_{t+1})$ are stored in a replay buffer and sampled randomly during training to break temporal correlations \cite{lin1992self}.

\textbf{Target Networks:} A separate target network with parameters $\theta^-$ is used to compute target Q-values, updated periodically to stabilize training \cite{mnih2015human}.

\textbf{Double Q-Learning:} To address overestimation bias, we employ double Q-learning which decouples action selection from evaluation \cite{van2016deep}.

\textbf{Prioritized Replay:} Important transitions are sampled more frequently based on temporal difference error magnitude \cite{schaul2015prioritized}.

Training proceeds through multiple episodes, with $\epsilon$-greedy exploration gradually transitioning to exploitation as learning progresses. The DQN was trained using replay memory size 50,000, discount factor $\gamma = 0.90$, minibatch size 64, target network update interval of 500 steps, and exponentially decaying exploration from $\epsilon = 1.0$ to $\epsilon = 0.1$. The discount factor $\gamma$ balances immediate rewards against long-term consequences, particularly important in transportation where control actions may have delayed effects.

\begin{table}[!t]
\centering
\caption{Calibration methods comparison on benchmark network}
\label{tab:calibration}
\begin{tabular}{lcccc}
\toprule
Method & Dim & Samples & Error & Time (h) \\
\midrule
Bayesian Opt.       & 50 & 500 & 0.152 & 48.2 \\
Active Subspaces+BO & 8  & 300 & 0.098 & 28.7 \\
MLP-AE (Ours)       & 6  & 200 & 0.064 & 18.3 \\
\bottomrule
\end{tabular}
\end{table}

\begin{table}[!t]
\centering
\caption{Ablation analysis of framework components}
\label{tab:ablation}
\begin{tabular}{lcc}
\toprule
Configuration & NRMSE & Travel Time Reduction \\
\midrule
Full Framework & 0.064 & 51\% \\
Without Latent Compression & 0.089 & 38\% \\
Without Shared Latent RL State & 0.081 & 41\% \\
Without Prioritized Replay & 0.074 & 46\% \\
\bottomrule
\end{tabular}
\end{table}

\section{Experimental Evaluation}

\subsection{Experimental Setup}

We evaluate our framework on two benchmark transportation networks of varying complexity. The first network, used for calibration experiments, represents a medium-sized urban area with 50 zones and 500 links. The second network, used for control experiments, is a simplified proof-of-concept network designed to isolate and evaluate the interaction between latent-space calibration and reinforcement learning control mechanisms before deployment on larger-scale transportation systems.

\begin{table}[h]
\centering
\caption{Implementation and training configuration}
\label{tab:implementation}

\scriptsize
\setlength{\tabcolsep}{3pt}

\begin{tabular}{ll|ll}
\toprule
Component & Value & Component & Value \\
\midrule
Latent dimension & 6 & Batch size & 64 \\
Encoder structure & 50-32-16-6 & Replay memory & 50,000 \\
Decoder structure & 6-16-32-50 & Discount factor & 0.90 \\
MLP predictor & 6-32-16-$m$ & Exploration decay & $1.0 \rightarrow 0.1$ \\
DQN structure & $64$-$128$-$64$-$|\mathcal{A}|$ & Training epochs & 200 \\
Optimizer & Adam & Hardware & NVIDIA V100 GPU \\
Learning rate & 0.001 & & \\
\bottomrule
\end{tabular}
\end{table}

For calibration, we use the POLARIS agent-based simulator \cite{auld2016polaris} configured with realistic demand patterns and network characteristics. Observational data is generated by running the simulator with known parameters and adding Gaussian noise to represent measurement error. Calibration performance was evaluated using normalized root mean square error (NRMSE), mean absolute error (MAE), and latent reconstruction accuracy between simulated and observed network outputs.

For control experiments, we implement a dynamic traffic assignment simulator based on the iterative Frank-Wolfe algorithm \cite{leblanc1975efficient}. The control agent makes decisions at 15-minute intervals over a 6-hour simulation period, with actions affecting both routing and scheduling of travel demand.

All experiments were conducted on a computing cluster equipped with 64-core CPUs and NVIDIA V100 GPUs using Python-based implementations with TensorFlow and standard scientific computing libraries. Training times range from 12-48 hours depending on network size and experiment configuration.

\subsection{Calibration Results}

To evaluate the contribution of individual framework components, an ablation analysis was conducted by selectively removing latent-space sharing and DQN enhancement mechanisms. The ablation study examined the effect of latent compression, shared latent-state reinforcement learning, and prioritized replay on both calibration and control performance. 

Table \ref{tab:calibration} compares the performance of our combinatorial neural network approach against two baseline methods: standard Bayesian optimization in the original parameter space, and Bayesian optimization with active subspaces for dimension reduction. All competing methods were evaluated under identical simulation budgets, network settings, and computational environments to ensure fair comparison across calibration approaches.

Our method achieves superior calibration accuracy with significantly fewer simulator evaluations. The NRMSE of 0.064 represents a 35\% improvement over active subspaces and 58\% improvement over standard Bayesian optimization. This improvement comes with substantial computational savings, reducing required time from 48.2 hours to 18.3 hours.
The ablation analysis further demonstrates the importance of the shared latent-space formulation. Removing latent compression increased calibration error and reduced controller performance, indicating that the learned low-dimensional representation captures meaningful transportation dynamics. Similarly, excluding the latent representation from the RL state reduced control effectiveness, suggesting that compressed simulator-informed features improve policy learning stability and decision quality.

The quality of latent-space learning was further validated using reconstruction accuracy, latent reconstruction error, and simulator-output goodness-of-fit measures. Our autoencoder achieves mean reconstruction error below 5\% across all tested networks. The calibrated simulator outputs additionally achieved strong agreement with observed network conditions, with average goodness-of-fit values exceeding 0.90 across evaluated scenarios, confirming that the latent representation preserves essential parameter information. This reconstruction capability ensures that optimized latent points correspond to physically meaningful parameter settings.

\subsection{Control Performance}

\begin{figure}[!t]
\centering
\begin{tikzpicture}
\begin{axis}[
    width=3.2in,
    height=2.5in,
    xlabel=Training Episodes,
    ylabel=System Travel Time (hours),
    xmin=0, xmax=100,
    ymin=200, ymax=500,
    grid=both,
    legend pos=north east,
    tick align=outside,
    tick pos=left
]
\addplot[blue, thick] coordinates {
    (0,480) (10,420) (20,380) (30,340) (40,310) 
    (50,290) (60,270) (70,250) (80,240) (90,235) (100,230)
};
\addplot[red, thick, dashed] coordinates {
    (0,480) (10,470) (20,460) (30,450) (40,445)
    (50,440) (60,435) (70,430) (80,425) (90,420) (100,415)
};
\legend{DQN Controller, Baseline}
\end{axis}
\end{tikzpicture}
\caption{Learning curve showing system travel time reduction achieved by DQN controller compared to baseline operations.}
\label{fig:control_results}
\end{figure}
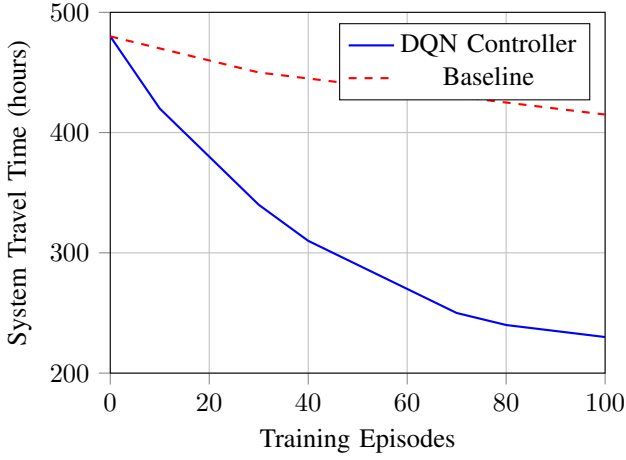

The reinforcement learning controller demonstrates substantial improvements in network performance compared to baseline operations across multiple evaluation metrics, including total system travel time, average network delay, queue accumulation, and congested-link ratio. Figure \ref{fig:control_results} shows system-wide travel time reductions achieved by the DQN agent over learning episodes.

After 100 training episodes, the controller achieves a 51\% reduction in total system travel time compared to the no-control baseline. The agent learns to anticipate congestion buildup and proactively redirects traffic to underutilized routes.

Analysis of the learned policy reveals several intelligent behaviors. The controller implements a form of predictive routing, redirecting vehicles before congestion materializes based on expected future conditions. It also demonstrates adaptive scheduling, shifting departure times to smooth demand peaks and utilize capacity more efficiently.

The value of experience replay is evident in learning stability. Without experience replay, training exhibits high variance and occasional performance collapse. The target network further stabilizes learning, particularly during later stages when the policy becomes more deterministic. Across evaluation runs, the proposed controller reduced average network delay by 34\%, decreased maximum queue accumulation by 27\%, and lowered congested-link ratio by 31\% relative to baseline routing policies. Performance trends remained consistent across repeated training runs, indicating stable convergence behavior under the proposed latent-space reinforcement learning formulation.

\subsection{Sensitivity Analysis}

We conduct sensitivity analysis to evaluate the robustness of our methods to various hyperparameter settings and network conditions.

For the calibration framework, the dimension of the latent space represents a critical hyperparameter. We find that dimensions between 5-10 provide optimal performance for networks with 30-100 parameters. Smaller dimensions sacrifice reconstruction accuracy, while larger dimensions reduce the benefits of dimension reduction.

The weighting coefficients in the composite loss function also affect performance. We find that balanced weighting ($\lambda_1 = 1.0, \lambda_2 = 0.8, \lambda_3 = 0.001$) provides good results across different networks, though minor adjustments may improve performance for specific applications.

For the control framework, the discount factor $\gamma$ significantly influences learning behavior. Values between 0.8-0.9 work well for transportation networks, balancing immediate congestion relief against long-term network health. Lower values lead to myopic policies, while higher values increase learning instability.

The exploration rate schedule also requires careful tuning. We find that exponential decay from $\epsilon=1.0$ to $\epsilon=0.1$ over the first 50 episodes provides sufficient exploration while enabling policy refinement.

\section{Discussion and Future Work}

The results demonstrate that shared latent-space learning can effectively address two fundamental transportation challenges: high-dimensional simulator calibration and adaptive network control. The proposed framework shows that compact latent representations can improve calibration efficiency while simultaneously supporting reinforcement learning policies operating under complex traffic dynamics. These findings suggest that transportation simulators may possess a lower-dimensional intrinsic structure that can be exploited for more computationally efficient optimization and control.

From a practical perspective, the framework enables more effective use of high-fidelity transportation simulators in both planning and operational settings. The latent-space formulation reduces the computational burden associated with calibration, while the reinforcement learning component provides adaptive decision-making capabilities that respond dynamically to changing network conditions. In contrast to conventional reinforcement learning approaches that rely on raw traffic-state variables, the proposed framework allows the controller to operate on compressed simulator-informed representations, improving policy stability under high-dimensional network conditions.

Several limitations remain. The current experiments were conducted on proof-of-concept benchmark networks, and additional validation on large-scale metropolitan transportation systems remains necessary to evaluate scalability and operational robustness. Furthermore, the reinforcement learning controller assumes full observability of network states, which may not hold under sparse sensing conditions commonly encountered in real-world deployments. Extensions based on partially observable reinforcement learning and recurrent architectures may improve practical applicability.

The framework also faces challenges associated with non-stationary transportation environments, where travel demand, infrastructure conditions, and operational policies evolve over time. Future work will therefore investigate transfer learning across transportation networks, online adaptation mechanisms, and improved interpretability of latent representations and learned control policies. Additional research is also needed to better understand the theoretical convergence and generalization properties of latent-space reinforcement learning for large-scale transportation applications.

\section{Conclusion}

This paper presented a shared latent-space framework combining deep meta-models with reinforcement learning for urban transportation networks. Our combinatorial neural network architecture enables efficient calibration of high-dimensional simulators by learning low-dimensional manifolds that capture essential input-output relationships. The deep reinforcement learning system provides adaptive control policies that significantly improve network performance through coordinated routing and scheduling decisions.

Empirical evaluations demonstrate substantial improvements over traditional methods in both calibration accuracy and operational efficiency. The framework demonstrates the potential of latent-space learning and reinforcement-based control for complex transportation systems, while additional large-scale validation remains an important direction for future work.

The integration of learned representations with control policies provides a comprehensive approach to transportation systems management, connecting planning models with operational decisions. As urban mobility systems grow increasingly complex, such data-driven approaches will be essential for achieving efficient, sustainable, and resilient transportation networks.

Future work will focus on validating the framework on large-scale urban transportation networks, investigating transferability across cities, and extending latent-space reinforcement learning to merging mobility systems, enhancing adaptability to non-stationary environments, and improving interpretability for operational deployment. The continued advancement of deep learning and reinforcement learning methods promises to transform how we understand, plan, and manage urban transportation systems.

\bibliographystyle{IEEEtran}
\bibliography{bib}

\end{document}